\documentclass[11pt]{article}
\usepackage{amssymb}
\usepackage[final]{acl}

\usepackage{times}
\usepackage{latexsym}
\usepackage{graphicx}
\usepackage{hyperref}

\usepackage{xcolor}
\usepackage[font=small]{caption}
\usepackage{float}
\usepackage{subcaption}
\usepackage{booktabs}
\usepackage{mwe}
\usepackage{amsmath}
\usepackage{bm}
\usepackage{array}
\usepackage{multirow}
\usepackage{longtable}
\usepackage{adjustbox}
\usepackage{tcolorbox}
\usepackage{diagbox}

\usepackage[T1]{fontenc}

\usepackage[utf8]{inputenc}

\usepackage{microtype}

\usepackage{inconsolata}

\usepackage{graphicx}

\title{%
  \raisebox{-0.3\height}{\includegraphics[height=0.9cm]{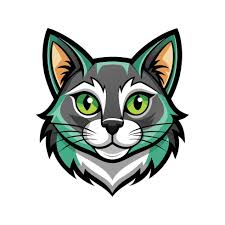}}\hspace{0.2em}%
CodeGENCAT: Generative Computerized Adaptive Testing for Open-ended Coding Problems}

\author{
\bf Wanyong Feng$^1$, Alexander Scarlatos$^1$, Ruochen Sun$^2$, Andrew Lan$^1$ \\
  University of Massachusetts Amherst$^1$, Independent Researcher$^2$ \\
  \texttt{wanyongfeng@umass.edu} \\
  }

\begin{document}
\maketitle
\begin{abstract}
Existing Computerized Adaptive Testing (CAT) frameworks typically select questions based on the predicted likelihood that the student will answer correctly. This design ignores information contained in students' open-ended responses, especially in domains such as programming education, where code structures and bugs contain rich information on student knowledge. In this work, we propose \textbf{Code} \textbf{GEN}erative \textbf{CAT} (\textbf{CodeGENCAT}), a generative CAT framework that selects questions using predicted student code responses. First, we develop a Generative Item Response Theory (GIRT) model that generates code responses conditioned on estimated student knowledge, trained with supervised fine-tuning followed by direct preference optimization for knowledge-response alignment. Second, we introduce three question-selection algorithms that measure uncertainty, coding style diversity, and information from predicted student code responses. Experiments on two real-world programming education datasets show that CodeGENCAT outperforms all CAT baselines, achieving an AUC improvement of up to 4.32\% over the strongest baseline in the early stages of adaptive testing.
\end{abstract}

\section{Introduction}
Computerized Adaptive Testing (CAT) is an important component of many standardized tests, such as the Graduate Record Examination (GRE), where questions are adaptively selected based on previous student responses and estimates of future performance \cite{luecht2011review}. 
A common strategy in CAT is to select a question that a student is roughly 50\% likely to answer correctly \cite{lord2012applications}, therefore maximizing the information gained from their answer; many systems use Item Response Theory (IRT) \cite{rasch1993probabilistic} to model the probability of a correct student answer.
However, IRT compresses each student response into a binary correctness label. While this abstraction is convenient for psychometric modeling, it discards rich diagnostic information in open-ended responses.

This limitation is especially important in programming education, where student responses are open-ended code submissions. Two students may both fail the same coding question, yet their responses can reflect different levels of understanding: one may make a small syntax error, while the other may not know how to answer the question at all. Treating both responses as simply incorrect ignores \textit{why} they are incorrect, which is important for identifying differences in conceptual understanding and coding ability.
Prior work in programming education has shown that open-ended code submissions enable accurate modeling of knowledge and learning \citep{shi2022code,duan2025automated}.

In this paper, we propose \textbf{Code} \textbf{GEN}erative \textbf{CAT} (\textbf{CodeGENCAT}), a generative CAT framework for programming education that selects questions based on open-ended student code responses rather than binary correctness. We first develop a Generative Item Response Theory (GIRT) model that uses estimated student knowledge to generate code responses to unseen coding questions. We train GIRT model in two stages: supervised fine-tuning to jointly learn student knowledge and response generation, followed by direct preference optimization to better align generated responses with estimated student knowledge. CodeGENCAT then selects questions using three criteria computed from the generated code responses: uncertainty in predicted correctness, coding style diversity, and information about the student's knowledge. We evaluate CodeGENCAT on two real-world coding education datasets: CodeWorkout, which contains Java submissions \citep{edwards2017codeworkout}, and ProgFeed, which contains Python submissions. Experiments show that CodeGENCAT improves knowledge estimation over numerous CAT baselines, particularly in the early stages of adaptive testing. Notably, on Codeworkout, our diversity-based selection algorithm achieves an AUC improvement of 4.32\% over the strongest CAT baseline at the first CAT stage, while our information-based selection algorithm consistently outperforms baselines across all test stages. \footnote{Our code is publicly available at \url{https://github.com/umass-ml4ed/GENCAT}}

\section{Background and Related Work}
A standard CAT framework has three components: a response model, a question bank, and a question-selection algorithm \cite{han2018components,van2000computerized}. The response model estimates the student's knowledge from observed responses. The question bank provides the set of candidate questions, usually with pre-estimated item properties such as difficulty. Given the current knowledge estimate, the question-selection algorithm chooses the next question expected to provide the most information about the student. This process repeats until a stopping criterion is met, such as reaching a fixed test length or sufficient confidence in the knowledge estimate. In most CAT settings, student responses are represented as binary correctness labels, and the goal is to estimate student knowledge accurately with as few questions as possible.

IRT is widely used as the response model in CAT \cite{van2009item}. In its simplest form, the one-parameter logistic (1PL) model, the probability that the student $i$ answers the question $j$ correctly is defined as $P(Y_{i,j}=1) = \sigma(\theta_i - b_j)$,
where $Y_{i,j} \in \{0,1\}$ is the response correctness, $\theta_i$ is the student's latent ability, $b_j$ is the question difficulty, and $\sigma(\cdot)$ is the sigmoid function. The 1PL model could be extended by introducing additional item parameters, such as discrimination in 2PL and guessing in 3PL, or by representing student knowledge as a vector in multidimensional IRT \cite{demars2010item,reckase2009historical}. Under 1PL IRT, question informativeness is commonly measured by Fisher information:
$I_j(\theta_i) = \sigma(\theta_i - b_j) \left(1 - \sigma(\theta_i - b_j)\right). $
The Fisher information is maximized when the predicted correctness probability is close to $0.5$, meaning the question is neither too easy nor too difficult given the current student ability estimate \cite{birnbaum1968some, chang1996global}.

Recent CAT frameworks have improved either the response model or the question-selection algorithm. BOBCAT formulates CAT as a bilevel optimization problem that jointly learns a response model and a question-selection algorithm \cite{ghosh2021bobcat}. NCAT learns a question-selection policy with reinforcement learning \cite{zhuang2022fully}. BECAT treats question selection as a data summarization problem to reduce student knowledge estimation error \cite{zhuang2023bounded}. GMOCAT uses a graph-enhanced multi-objective formulation for question selection \cite{wang2023gmocat}. LACAT uses LLM agents to improve the interpretability of question selection \cite{cheng2024towards}. Despite these advances, a critical limitation in CAT remains in how it utilizes question and response data. With the exception of LACAT, which utilizes question text for reasoning, most existing CAT frameworks simply represent questions as scalar-valued parameters (difficulty, etc.), discarding the rich textual information contained in the question. However, even LACAT reduces the student response to a binary correctness value.

There have been works that model student knowledge in programming education. Several works modeled dynamic student knowledge via knowledge tracing to predict open-ended student code \cite{liu-etal-2022-open, duan2025test, duan2025automated}, binary success in programming exercises \cite{wang2017learning}, or high-level coding errors \cite{shaka-etal-2024-error}. Other works modeled open-ended student code using static knowledge representations, and align models with signals based on ability \cite{scarlatos2025smart} and coding errors \cite{duan2026kaser}. Many of these works are supported by pretrained code models \cite{feng2020codebert,chen2021evaluating}.


\section{Methodology}
In this section, we first formally define the data format, and then detail the components of our CodeGENCAT framework.
\begin{figure*}[t]
\vspace{-1cm} 
\centering
\includegraphics[width=.9\linewidth]
{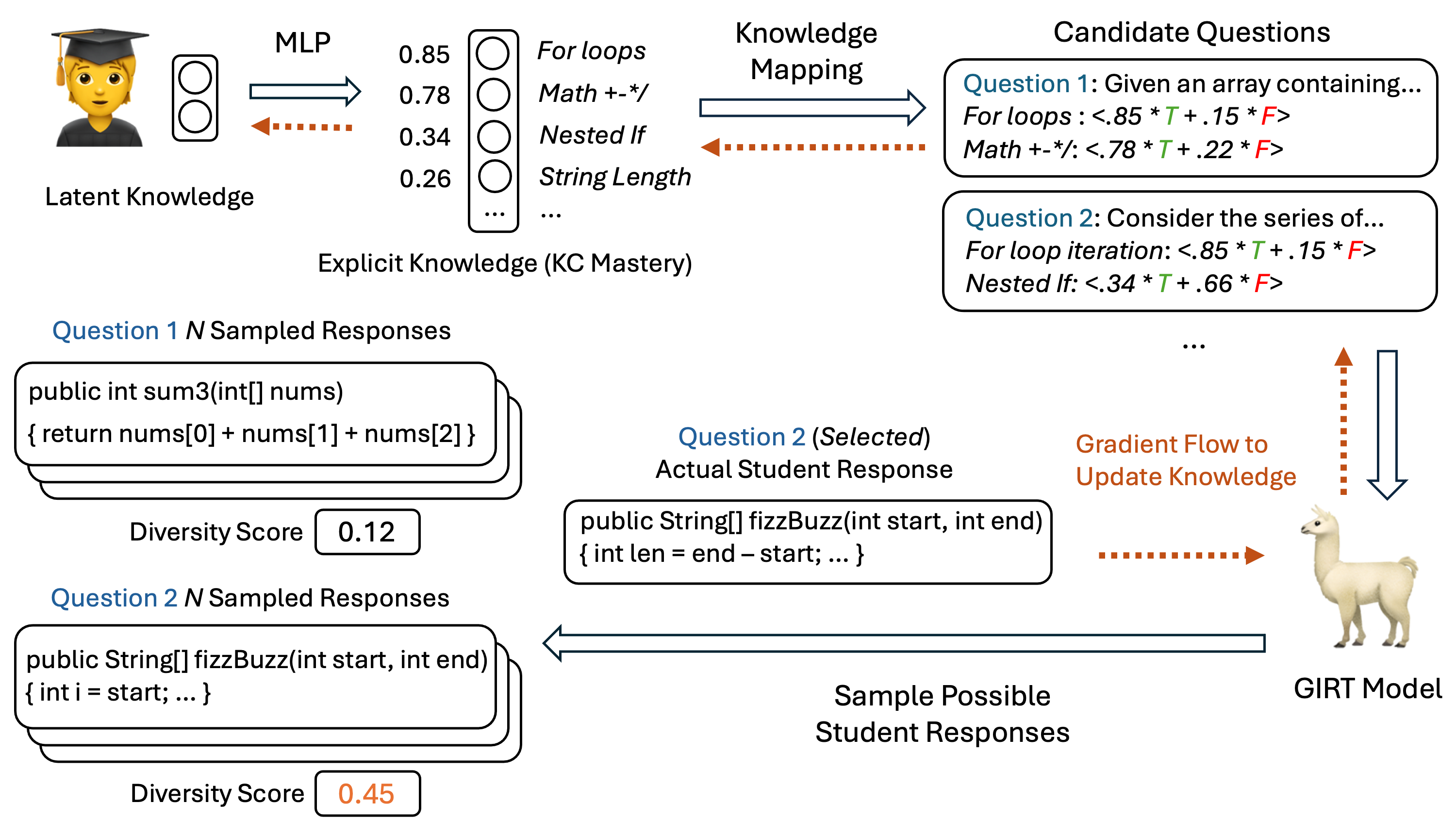}
\caption{Overview of one CodeGENCAT cycle with Diversity as the question selection algorithm.}
\label{fig:training_process}
\vspace{-0.5cm}
\end{figure*}

\subsection{Data Format}
Let $\mathcal{U} = \{u_1, u_2, \dots, u_I\}$ denote the set of $I$ students and $\mathcal{Q} = \{q_1, q_2, \dots, q_J\}$ denote the set of $J$ coding questions. We define the set of distinct knowledge components (KCs) 
as $\mathcal{K} = \{k_1, k_2, \dots, k_M\}$. A KC represents a fundamental knowledge/skill required to answer a coding question (e.g., loops, arrays, and arithmetic operations). Each coding question $q_j \in \mathcal{Q}$ is denoted as a tuple $(p_j, \mathcal{K}_j)$, where $p_j$ is the question text sequence and $\mathcal{K}_j \subseteq \mathcal{K}$ is the subset of KCs associated with the question. We define the dataset of student responses as $\mathcal{R} = \{x_{i,j} \mid (i,j) \in \Omega\}$, where $\Omega \subseteq \{1 \dots I\} \times \{1 \dots J\}$ represents the set of observed student responses; not every student necessarily responds to every question. Each data point $x_{i,j}$ corresponds to student $u_i$'s response to question $q_j$, defined as:
$x_{i,j} = (c_{i,j}, a_{i,j}, q_j)$,
where $c_{i,j}$ is the student's response, which is a sequence of code tokens, and $a_{i,j} \in [0,1]$ is the response correctness, which represents the portion of unit test cases that $c_{i,j}$ passed.

\subsection{Generative Item Response Theory Model}
We now detail our model for generating ``student'' code based on estimated knowledge. There are two key components to our model: 1) latent knowledge representation, and 2) LLM-driven code generation. Inspired by \cite{duan2025automated}, we represent the explicit knowledge of student $u_i$ as a vector of mastery levels across all $M$ distinct KCs, denoted as $\boldsymbol{\theta}_i = (\theta_{i,1}, \theta_{i,2}, \dots, \theta_{i,M})$, where $\theta_{i,m} \in [0,1]$ is 1 if the student has full mastery of the KC and 0 if they have no mastery of it.
While we can learn each $\theta_{i,m}$ independently, doing so ignores interdependencies between KCs (e.g., mastery of arrays often implies mastery of loops).
Therefore, we introduce a low-dimensional latent knowledge representation $\mathbf{z}_i \in \mathbb{R}^D$ ($D \ll M$).
We project this latent knowledge into the explicit knowledge space using a multilayer perceptron (MLP) parameterized by $\epsilon$, followed by a sigmoid activation to constrain the mastery levels to the unit interval: $\boldsymbol{\theta}_i = \sigma(\mathrm{MLP}_\epsilon(\mathbf{z}_i)).$

We use an LLM parameterized by $\phi$ to predict a student's code response to a question, given their current knowledge estimate: $P_\phi(c_{i,j} \mid q_j, \boldsymbol{\theta}_i)$. 
To enable joint learning of the student latent knowledge $\mathbf{z}_i$, the projection parameters $\epsilon$, and the LLM parameters $\phi$, we introduce a differentiable transformation $g: [0,1] \to \mathbb{R}^H$, which maps a scalar mastery level to a continuous embedding vector compatible with the LLM's hidden dimension $H$.
Specifically, for each KC $k \in \mathcal{K}_j$, we compute a mastery embedding $\mathbf{e}_{i,k} = g(\theta_{i,k})$. In our implementation, $g(\cdot)$ is a linear interpolation between the pre-trained embeddings of the semantic tokens ``TRUE'' and ``FALSE'':
\begin{equation*}
\mathbf{e}_{i,k} = \theta_{i,k} \cdot \mathbf{e}^{\text{TRUE}} + (1 - \theta_{i,k}) \cdot \mathbf{e}^{\text{FALSE}}. 
\end{equation*}
This soft prompting approach enables the LLM to leverage its pre-trained semantic understanding to represent KC mastery. A higher mastery value $\theta_{i,k}$ shifts the representation closer to ``TRUE'', signaling higher master level of KC $k$. We inject these embeddings into the input prompt using a structured template: \textit{``KC 1: <$k$>. The student has knowledge of <$k$>: <$\mathbf{e}_{i,k}$>...''}. The LLM then generates the predicted code response conditioned on this knowledge-augmented prompt. See one specific example in Figure~\ref{fig:training_process}. See Appendix~\ref{tab:gir_model_prompt} for the exact prompt we used.

\subsection{GIRT Model Training}
We first leverage supervised fine-tuning (SFT), where we define the generative loss for each $x_{i,j}$ as the negative log-likelihood of the student response $c_{i,j}$ given the question text and the student latent knowledge:
$\mathcal{L}_{i,j}^{\text{SFT}} = -\log P_{\phi, \epsilon}(c_{i,j} \mid q_j, \mathbf{z}_i)$.

However, SFT alone does not guarantee that the explicit knowledge $\boldsymbol{\theta}_i$ is aligned with student correctness.
To ensure that knowledge estimates are interpretable, we need to enforce monotonicity, i.e., higher KC mastery leads to higher probability of a correct response.
Therefore, we introduce an auxiliary alignment loss to align student explicit knowledge with correctness of the open-ended response, $a_{i,j}$.
We first take the average knowledge of the student on the associated KCs, $\mathcal{K}_j$, and use it as the predicted correctness, i.e., $\hat{a}_{i,j} = \frac{1}{|\mathcal{K}_j|} \sum_{k \in \mathcal{K}_j} \theta_{i,k}.$
We then apply a binary cross-entropy loss between this predicted success probability and the ground-truth correctness score: $\mathcal{L}_{i,j}^{\text{KC}} = -\bigl[ a_{i,j} \log \hat{a}_{i,j} + (1 - a_{i,j}) \log(1 - \hat{a}_{i,j}) \bigr]$,
for each data point $x_{i,j}$. We note that $a_{i,j}$ can either be continuous or binary-valued, i.e., $a_{i,j}=1$ if the student code passes all test cases and $a_{i,j}=0$ otherwise. Although this choice does not affect the binary cross-entropy loss, we follow standard IRT literature and use $a_{i,j} \in \{0,1\}$ in what follows. 
Finally, we combine the two losses using a hyperparameter $\lambda \in [0,1]$. The total loss for a single data sample is: $\mathcal{L}_{i,j} = (1-\lambda)\mathcal{L}_{i,j}^{\text{SFT}} + \lambda\mathcal{L}_{i,j}^{\text{KC}}.$
Finally, the global training objective is the average loss over the entire set of observed data samples $\mathcal{R}$: $\mathcal{L}_{\text{Total}}(z, \epsilon, \phi) = \frac{1}{|\mathcal{R}|} \sum_{x_{i,j} \in \mathcal{R}} \mathcal{L}_{i,j}.$
We train the student latent knowledge ($z$), MLP ($\epsilon$), and LLM ($\phi$) end-to-end by optimizing these parameters together to minimize $\mathcal{L}_{\text{Total}}$.

\subsection{Knowledge-Response Alignment}
While our GIRT model is designed to align generated responses with conditioned knowledge, we observe that SFT alone does not reliably generate responses that align with the knowledge in the prompt. Specifically, the post-SFT model suffers from mode collapse, generating semantically identical responses for some questions regardless of the input student knowledge.
In other words, the GIRT model often ignores the input student knowledge, failing to mimic errors made by struggling students or capture diversity among correct responses.

To address this shortcoming, we introduce a reinforcement learning stage where we reward alignment between response correctness and input student knowledge. Specifically, following \cite{scarlatos2025smart}, we define a reward function using the likelihood of a response given student latent knowledge, form preference pairs among responses in the data, and refine the GIRT model using direct preference optimization (DPO) \cite{rafailov2023direct}.

We define our reward function, corresponding to the likelihood of a response given student knowledge, using a continuous Bernoulli distribution \cite{loginova2021towards}:
$P(a_{i,j} \mid \mathbf{z}_{i}) = \hat{a}_{i,j}^{a_{i,j}} \cdot (1 - \hat{a}_{i,j})^{1-a_{i,j}}$.
To form preference pairs for a specific data sample $x_{i,j}$, we pair the actual response $c_{i,j}$ (preferred) with a response $c_{i',j}$ from another student $u_{i'}$ (dispreferred) if it is less likely to be responded by student $u_i$ by a threshold $\tau$. Specifically, $c_{i',j}$ is a valid dispreferred candidate if it satisfies
$P(a_{i,j} \mid \mathbf{z}_{i}) - P(a_{i',j} \mid \mathbf{z}_{i}) > \tau$.
We take the $M$ candidate $c_{i',j}$ responses with the highest likelihood differences to form $M$ preference pairs for DPO. We keep knowledge estimates frozen during DPO to ensure a stable alignment target.

\subsection{CAT Cycles}
CAT represents an iterative process that alternates between selecting the most informative question and updating the student knowledge estimate. We detail our iterative process while focusing on a single student $u_i$. Let $\mathcal{Q}_i \subseteq \mathcal{Q}$ denote the set of candidate questions available to the student. Each CAT time step consists of three phases:

\paragraph{Initialization} To warm-start the CAT iterations before observing any responses for a new student, we initialize their latent knowledge $\mathbf{z}_i^{(0)}$ using a population prior. Specifically, we set this prior to the average of latent knowledge vectors across all students from GIRT training.

\paragraph{Response Sampling} At each time step $t$, we provide a score to each of the available candidate questions $\mathcal{Q}_{i,t}$, where the highest scoring one will be given to the student next. For each candidate question $q_j \in \mathcal{Q}_{i,t}$, we first sample $N$ responses from the GIRT model, conditioned on the latent knowledge estimate from the previous time step:
$\hat{c}_{i,j}^{(n)} \sim P_{\phi, \epsilon}(\cdot \mid q_j, \mathbf{z}_i^{(t-1)})$.
Then, we apply one of our question selection algorithms (detailed in Section~\ref{sec:algo}) to these sampled responses, resulting in a score used to select the next question, $q_{j^*}$. 

\paragraph{Knowledge Update} After observing the student's actual response to $c_{i,j^*}$ to $q_{j^*}$, we update their latent knowledge $\mathbf{z}_i$ while keeping the GIRT model parameters ($\phi, \epsilon$) fixed. We update $\mathbf{z}_i$ via gradient descent to minimize the total loss across the student responses to all observed questions.
We visualize the iterative process in Figure~\ref{fig:training_process}.

\subsection{Question Selection Algorithms}
\label{sec:algo}
We explore three algorithms to select the next question among all candidates, each using a different way to evaluate question informativeness. We define a scoring metric for each algorithm, and select the question with the maximum score as the next one to give to the student. 

\paragraph{Uncertainty-Based} This algorithm selects the question with the maximum response correctness prediction uncertainty. Inspired by the informativeness measure in the traditional 1PL CAT selection algorithm, we assume that the question with a predicted correctness closest to 0.5 will provide the most information on student knowledge. For a candidate question $q_j$, we first evaluate the $N$ sampled responses using a trained scoring model, which takes the question and response as input and outputs a continuous correctness score $\hat{a}_{i,j}^{(n)} \in [0,1]$. We provide details on the scoring model in Appendix~\ref{app:score}.
We then construct the uncertainty metric $U_j$, which encourages average scores close to 0.5:
\begin{equation*}
\textstyle U_j = 1 - \left| \left( \frac{1}{N} \sum_{n=1}^N \hat{a}_{i,j}^{(n)} \right) - 0.5 \right|.
\end{equation*}

\paragraph{Diversity-Based} This algorithm selects the question with the most semantically diverse set of generated responses. High semantic diversity indicates that the GIRT model is unsure how the student might respond, reflecting a high degree of uncertainty. For a candidate question $q_j$, we first use a pre-trained \textbf{CodeBERT} model \cite{feng2020codebert} to obtain code semantic embedding vectors $\{\mathbf{h}_{i,j}^{(n)}\}_{n=1}^N$ for the $N$ sampled responses. We then construct the diversity metric $S_j$, which encourages low pairwise cosine similarities across sample embeddings:
\begin{equation*}
\textstyle S_j \!= 1 - \left[ \frac{2}{N(N-1)} \sum_{n=1}^N \sum_{m=n+1}^N \!
\frac{\mathbf{h}_{i,j}^{(n)} \cdot \mathbf{h}_{i,j}^{(m)}}{\lvert \lvert \mathbf{h}_{i,j}^{(n)} \rvert \rvert \, \lvert \lvert \mathbf{h}_{i,j}^{(m)} \rvert \rvert}
\right].
\end{equation*}

\paragraph{Fisher-Information-Based} This algorithm selects the question with the maximum Fisher information. Specifically, we use the trace of the Fisher information matrix as the scalar information score \cite{pukelsheim2006optimal}. High Fisher information indicates that observing the response is expected to substantially reduce uncertainty about the student's latent knowledge. For a candidate question $q_j$, we compute this score with respect to the student's latent knowledge vector $\mathbf{z}_i$ across $N$ sampled responses. Since the corresponding Fisher information matrix is the outer product of the log-likelihood gradient and its transpose, its trace is equivalent to the squared $\ell_2$ norm of the gradient. We construct the Fisher information metric $\mathcal{I}_j$ as:
\begin{equation*}
\textstyle \mathcal{I}_j \!=\! \frac{1}{N} \sum_{n=1}^N \left\|\nabla_{\mathbf{z}_i}\log P_{\phi, \epsilon}\left(\hat{c}_{i,j}^{(n)}\mid q_j,\mathbf{z}_i^{(t-1)}\right)\right\|_2^{2}.
\end{equation*}

 

\section{Experimental Evaluation}
We now detail our experiments to validate our methodology, where we leverage two real-world datasets and compare CodeGENCAT against several strong CAT baselines.

\subsection{Datasets Details}
The first dataset is \textbf{CodeWorkout}~\cite{edwards2017codeworkout}, which was originally released as part of the Second CSEDM Data Challenge \cite{csedm_dc_2021}. This dataset consists of Java code submissions collected from students in an introductory computer science course. Each question in the dataset is associated with a set of human-written KCs, with 18 distinct KCs in total. The dataset comprises 246 students and 50 programming questions. Following prior work \cite{shi2022code,duan2025automated}, we only analyze each student's first submission to each problem, since students are likely to eventually solve a problem correctly after subsequent attempts. After this filtering, the final dataset contains 10,834 student submissions in total. 

The second dataset is \textbf{ProgFeed}, which we collected ourselves. This dataset consists of Python code submissions collected from students in an introductory computer science course. Each question in the dataset is associated with a set of human-written KCs, with 10 distinct KCs in total. The dataset comprises of 374 students, 42 programming questions, and 8,724 student submissions in total. Similar to CodeWorkout, we only consider each student's first submission to a problem. We are in the process of undergoing IRB review to make ProgFeed publicly available.

For data splitting, we use 5-fold cross-validation to ensure robust results. In each fold, we randomly partition students into a GIRT model training group (80\%) and a CAT evaluation group (20\%). For students in the GIRT model training group, we further split their question responses into training, validation, and testing sets with a ratio of 8:1:1 for CodeWorkout, and into training and testing with a ratio of 7:3 for ProgFeed to ensure the test set contains enough student responses for meaningful evaluation. For students in the CAT evaluation group, we hold out a test set of 10 questions for CodeWorkout, and 5 questions for ProgFeed to evaluate whether the estimated student knowledge can accurately predict their response, following \cite{ghosh2021bobcat}. The remaining questions form the candidate question set for selection. The size of the candidate question set is at least 30 for CodeWorkout and 20 for ProgFeed. We provide further details on the data splitting in Appendix~\ref{app:data}.

\subsection{Baselines} We compare the three variants of CodeGENCAT (\textbf{Uncertainty}, \textbf{Diversity}, and \textbf{Information}) against four representative CAT baselines:
\textbf{1PL IRT}: The standard baseline in the CAT literature. It models the probability of a correct response under the 1PL IRT model and selects the next question with maximum Fisher Information. 
\textbf{BOBCAT} \cite{ghosh2021bobcat}: A framework that jointly learns both the response model and the question selection algorithm and uses bilevel optimization to minimize prediction error on the test set. 
\textbf{NCAT} \cite{zhuang2022fully}: A follow-up to BOBCAT that use reinforcement learning to learn a policy that maximizes prediction accuracy on the test set. 
\textbf{LACAT} \cite{cheng2024towards}: An LLM agent-based framework that uses multiple agents to decide which question to select next. Unlike CodeGENCAT, LACAT still relies on binary response correctness rather than open-ended responses.

\subsection{Experimental Setup and Metrics}
We use \textbf{Llama-3.2-1B-Instruct}  \cite{dubey2024llama} as the backbone for our GIRT model, and use a 3-layer MLP with ReLU activations for projecting latent student knowledge into explicit knowledge (KC mastery levels). We provide full training details in Appendix~\ref{app:exp}.

The goal of CAT is to accurately estimate student knowledge using as few questions as possible. However, in real-world data, a student's true knowledge state is not directly available. Therefore, we use the estimated student knowledge to predict their responses on the test set, which acts as a proxy for knowledge estimate accuracy. We use area under the receiver operating characteristic curve (\textbf{AUC}) and \textbf{Accuracy} to measure how well each method predicts the correctness of student responses.
To predict correctness with CodeGENCAT, we leverage the pre-trained 1PL IRT model: given the question IDs selected by CodeGENCAT and the corresponding ground-truth student responses, we fit an ability for the student and predict the probability of correct responses to test set questions using the IRT model.
Doing so provides continuous correctness probabilities from the method, and is more reliable than directly using the GIRT model's knowledge estimates, $\hat{a}_{i,j}$, to predict correctness.
%
We also evaluate the similarity of generated student code to ground-truth code on the held-out test set. We provide the details for this evaluation in Appendix \ref{app:codemetric}.

\section{Results, Analysis, and Discussion}
In this section, we detail quantitative results comparing CodeGENCAT with baselines on knowledge estimate and test security. We then conduct various ablation studies and qualitative analyses to see how different settings affect CodeGENCAT. 

\begin{table*}[t]
\centering
\resizebox{\textwidth}{!}{
\begin{tabular}{l ccccccc c ccccccc}
\toprule
\multirow{3}{*}{\textbf{Framework}} & \multicolumn{15}{c}{\textbf{Step $t$}} \\
\cmidrule(lr){2-16}
& \multicolumn{7}{c}{\textbf{CodeWorkout}} && \multicolumn{7}{c}{\textbf{ProgFeed}} \\
\cmidrule(lr){2-8} \cmidrule(lr){10-16}
& $1$ & $3$ & $5$ & $7$ & $10$ & $15$ & $20$ && $1$ & $3$ & $5$ & $7$ & $10$ & $15$ & $20$ \\
\midrule
\multicolumn{16}{c}{\textbf{AUC $\uparrow$}} \\
\midrule
CodeGENCAT (Uncertainty) & 0.6737 & $0.7258^*$ & 0.7415 & 0.7599 & 0.7693 & \underline{0.7838} & 0.7873 && 0.6375 & 0.6725 & \underline{0.6875} & \underline{0.6930} & 0.7020 & 0.7063 & 0.7101 \\
CodeGENCAT (Diversity)   & \textbf{0.7073} & $\textbf{0.7352}^*$ & $\textbf{0.7529}^*$ & $\textbf{0.7632}^*$ & 0.7687 & 0.7801 & 0.7855 && 0.6437 & 0.6319 & 0.6834 & 0.6870 & 0.6901 & 0.7064 & 0.7161 \\
CodeGENCAT (Information) & \underline{0.6854} & \underline{0.7264} & \underline{0.7492} & \underline{0.7616} & \textbf{0.7718} & \textbf{0.7842} & \textbf{0.7901} && \textbf{0.6730} & \textbf{0.6845} & \textbf{0.6910} & \textbf{0.7080} & \textbf{0.7056} & \textbf{0.7204} & 0.7191 \\
1PL IRT             & 0.6641 & 0.7039 & 0.7299 & 0.7474 & \underline{0.7701} & 0.7832 & \underline{0.7881} && 0.6545 & 0.6649 & 0.6680 & 0.6731 & 0.6867 & \underline{0.7132} & 0.7199 \\
LACAT                & 0.6287 & 0.6872 & 0.7104 & 0.7501 & 0.7605 & 0.7787 & 0.7842 && 0.6660 & 0.6576 & 0.6744 & 0.6805 & 0.6891 & 0.7072 & \textbf{0.7255} \\
NCAT                 & 0.6580 & 0.7110 & 0.7343 & 0.7470 & 0.7571 & 0.7689 & 0.7764 && \underline{0.6673} & \underline{0.6796} & 0.6741 & 0.6897 & \underline{0.7046} & 0.7130 & \underline{0.7231} \\
BOBCAT               & 0.5893 & 0.6845 & 0.7179 & 0.7288 & 0.7445 & 0.7495 & 0.7513 && 0.5193 & 0.5243 & 0.5479 & 0.5475 & 0.5566 & 0.5660 & 0.5725 \\
\midrule
\multicolumn{16}{c}{\textbf{Accuracy $\uparrow$}} \\
\midrule
CodeGENCAT (Uncertainty) & 0.6012 & \textbf{0.6796} & 0.6788 & \textbf{0.6980} & 0.6988 & \textbf{0.7148} & 0.7212 && 0.6296 & 0.7248 & 0.7448 & 0.7432 & 0.7616 & 0.7560 & 0.7656 \\
CodeGENCAT (Diversity)   & \textbf{0.6452} & 0.6720 & $\textbf{0.6888}^*$ & 0.6892 & 0.6984 & 0.7020 & 0.7136 && 0.6856 & 0.7176 & 0.7424 & 0.7496 & 0.7600 & 0.7704 & \textbf{0.7728} \\
CodeGENCAT (Information) & \underline{0.6328} & \underline{0.6740} & \underline{0.6876} & \underline{0.6964} & \textbf{0.7044} & 0.7068 & \textbf{0.7232} && 0.6960 & 0.7184 & \underline{0.7456} & \underline{0.7648} & \textbf{0.7648} & \textbf{0.7760} & \textbf{0.7728} \\
1PL IRT             & 0.5988 & 0.6552 & 0.6780 & 0.6808 & \underline{0.7036} & 0.7064 & \underline{0.7216} && 0.5904 & 0.7232 & 0.7376 & 0.7440 & 0.7544 & 0.7640 & \underline{0.7720} \\
LACAT                & 0.5620 & 0.6376 & 0.6536 & 0.6836 & 0.6896 & 0.7048 & 0.7120 && \underline{0.6976} & \textbf{0.7424} & \textbf{0.7536} & \textbf{0.7680} & \underline{0.7640} & \underline{0.7728} & 0.7712 \\
NCAT                 & 0.6168 & 0.6632 & 0.6728 & 0.6920 & 0.7012 & \underline{0.7128} & 0.7144 && \textbf{0.7240} & \underline{0.7288} & 0.7384 & 0.7504 & 0.7592 & 0.7584 & 0.7624 \\
BOBCAT               & 0.5220 & 0.5220 & 0.6380 & 0.6668 & 0.6804 & 0.6948 & 0.6900 && 0.2336 & 0.2336 & 0.6144 & 0.6760 & 0.7144 & 0.7336 & 0.7184 \\
\bottomrule
\end{tabular}
}
\caption{5-fold averaged knowledge estimate performance (AUC and Accuracy) on both datasets across varying test lengths ($t$). CodeGENCAT consistently outperforms baselines, particularly in the early stages ($t=1$ to $7$), with Diversity, and late stages ($t=10$ to $20$) with Information on the CodeWorkout dataset. CodeGENCAT consistently outperforms baselines for most time steps with Information on the ProgFeed dataset. * denotes statistically significant improvement over the best baseline using a paired sample t-test (p < 0.05). For AUC, the best baselines are 1PL IRT and NCAT, and for accuracy, the best baselines are NCAT and LACAT, for CodeWorkout and ProgFeed respectively. Best results are in \textbf{bold} and second best are \underline{underlined}.}
\label{tab:main_results}
\end{table*}

\subsection{Quantitative Results}
\textbf{CodeGENCAT excels in early-stage knowledge estimation.} We report the 5-fold averaged knowledge estimate performance (AUC and Accuracy) of CodeGENCAT and four baselines across test lengths ranging from $t=1$ to $t=20$ in Table~\ref{tab:main_results}. We observe that CodeGENCAT consistently yields more accurate knowledge estimates than all baselines, particularly during the early, most important phase of CAT ($t$ from 1 to 7). For example, CodeGENCAT outperforms the strongest baseline at $t=5$ on both CodeWorkout (Diversity: 0.7529 vs. NCAT: 0.7343) and ProgFeed (Information: 0.6910 vs. LACAT: 0.6744) on AUC.
This performance improvement stems from CodeGENCAT's ability to leverage both the question text and student-written code to select questions, shifting the focus from predicting \textit{whether} a student will fail to \textit{how} they might fail. By evaluating these sampled responses, CodeGENCAT captures valuable information about student knowledge, such as specific errors or misconceptions the student might have. As the test progresses and the knowledge estimate stabilizes in the later time steps, CodeGENCAT remains competitive, with the performance gap naturally narrowing since all CAT methods converge to the same performance when a student responds to every question.

\textbf{Performance varies with selection algorithm across test lengths.} Comparing our three question selection algorithms, we observe that the best selection algorithm depends on both the dataset and the stage of the test. On CodeWorkout, Diversity performs best when $t$ is small; for example, Diversity outperforms Information at $t=3$ (0.7352 vs. 0.7264) on AUC. This difference suggests that maximizing response diversity is effective when the student knowledge estimate does not diverge much from its initialized prior value ($\mathbf{z}_i^{(0)}$). Conversely, Information achieves better performance when $t$ increases; for example, Information outperforms Diversity at $t=15$ (0.7842 vs. 0.7801) on AUC. This difference suggests that when the student knowledge estimate becomes more accurate, gradient-based information is better suited for refining the knowledge estimate in the later phase of the test. On ProgFeed, Information is always the best selection algorithm. The potential reason could be the GIRT model has limited ability to generate diversified code given the questions in ProgFeed are simpler and more straightforward than those in CodeWorkout. Finally, Uncertainty generally yields lower AUC values than Diversity or Information on both datasets, suggesting that relying only on the predicted correctness of simulated code is insufficient for optimal question selection and further emphasizing the importance of explicitly modeling open-ended student responses.

\textbf{Baselines are limited by data sparsity and knowledge representation abstraction.} 
BOBCAT has the lowest performance among all baselines, likely because its bilevel optimization objective is data-hungry and may not work well when the question bank is small (50 in CodeWorkout compared to at least 1000 in datasets it was evaluated on). While LACAT uses strong LLM agents (\textbf{GPT-3.5-turbo-16k}) for question selection, it underperforms CodeGENCAT at $t=5$ on both CodeWorkout (0.7104 vs. 0.7529) and ProgFeed (0.6744 vs. 0.6910) on AUC. This gap suggests a critical limitation in LACAT's input student knowledge representation: it relies on the abstract knowledge parameter $\theta$ from 1PL to guide its agents to select questions. In contrast, CodeGENCAT represents student knowledge as mastery of granular KCs, which is more specific and interpretable to LLMs.

\begin{table}[t]
\centering
\resizebox{\columnwidth}{!}{%
\begin{tabular}{lcc}
\toprule
\textbf{Method} & \textbf{Exp.\% (med) $\downarrow$} & \textbf{Over.\% (mean) $\downarrow$} \\
\midrule
CodeGENCAT (Uncertainty)       & 12.80          & \textbf{25.33} \\
CodeGENCAT (Diversity)   & 11.60          & 36.21          \\
CodeGENCAT (Information)        & 15.60          & 35.41          \\
1PL IRT             & \underline{10.60}          & 28.20          \\
LACAT                & \textbf{6.40}  & \underline{25.92}          \\
NCAT                 & 41.40          & 58.92          \\
BOBCAT               & 12.60          & 35.67          \\

\bottomrule
\end{tabular}%
}
\caption{5-fold averaged median Exposure rate (Exp.) and mean overlap rate (Over.) at step $t = 5$ on CodeWorkout dataset. Best results in \textbf{bold} and second best are \underline{underlined}.}
\label{tab:exposure_overlap}
\end{table}

\paragraph{CodeGENCAT satisfies test security requirements.}
Enhancing test security, i.e., preventing item overexposure, leakage and ensuring the validity of the item pool and test results, is very important in CAT. Therefore, we also analyze the 5-fold averaged Question Exposure \textbf{(Exp)} and Test Overlap \textbf{(Over)} rates, which are common metrics for test security, at step $t=5$ for all question selection algorithms on CodeWorkout dataset. 
As shown in Table \ref{tab:exposure_overlap}, CodeGENCAT (Uncertainty) achieves the lowest test overlap rate of 25.33\%, indicating that on average, 25.33\% of the questions received by two different students are same. In terms of question exposure rate, while LACAT achieves the lowest median exposure rate (6.40\%), indicating that half of the questions are administered to at most 6.40\% of the students, CodeGENCAT remains competitive with other baselines, with CodeGENCAT (Diversity) and CodeGENCAT (Uncertainty) maintaining low exposure rates of 11.60\% and 12.80\% respectively. These results demonstrate that CodeGENCAT is capable of balancing test efficiency and security.

\subsection{Ablation Study}
\paragraph{Effect of Backbone Model Size.}
We investigate the effect of the GIRT backbone LLM size by comparing \textbf{Llama-3.2-1B-Instruct} and \textbf{Llama-3.2-3B-Instruct} in Table~\ref{tab:gencat_1b_3b_ablation}. On CodeWorkout, the 3B model generally improves AUC across selection algorithms; for example, 3B outperforms 1B for Diversity at $t=1$ (0.7235 vs. 0.7073) and for Information at $t=5$ (0.7655 vs. 0.7492). However, on ProgFeed, scaling the backbone does not consistently improve performance: 1B Information outperforms 3B Information at $t=1$ (0.6730 vs. 0.6400) and $t=3$ (0.6845 vs. 0.6551) on AUC, while 3B Information only slightly improves at $t=10$ and $t=15$. These results show that the benefit of a larger LLM is dataset-dependent. We use the 3B model for the remaining CodeWorkout ablations,  where scaling the backbone LLM provides clearer and more consistent gains.

\paragraph{Necessity of DPO Alignment.}
We investigate the necessity of DPO alignment in Table~\ref{tab:codeworkout_3b_dpo_sft_ablation}. For Diversity, DPO consistently improves AUC across all time steps; for example, DPO outperforms SFT at $t=1$ (0.7235 vs. 0.7079), and $t=5$ (0.7547 vs. 0.7444). For Information, DPO excels in the key early time steps, such as $t=3$ (0.7476 vs. 0.7380) and $t=5$ (0.7655 vs. 0.7616), but SFT performs better on later time steps. SFT also outperforms DPO for Uncertainty. This result suggests that aligning generated code with student knowledge increases diversity of generated code and improves the accuracy of response likelihood estimates used by Information. However, this alignment also pushes the predicted correctness scores toward 0 or 1, making it harder for Uncertainty to identify the best question.



\paragraph{Number of Samples.}
We investigate the impact of the number of sampled responses ($N$) in Table~\ref{tab:codeworkout_sample_ablation} by comparing $N=3$ and $N=10$ for each candidate question. Using $N=10$ achieves better performance at many time steps, but the improvements are usually small. For Diversity, $N=10$ outperforms $N=3$ at $t=1$ (0.7235 vs. 0.6927) and $t=10$ (0.7777 vs. 0.7746) on AUC. For Information, $N=10$ outperforms $N=3$ at $t=5$ (0.7655 vs. 0.7612) and $t=7$ (0.7717 vs. 0.7652). These small improvements show a clear trade-off between knowledge estimate accuracy and question selection efficiency. Reducing $N$ from 10 to 3 lowers the total adaptive testing time by at most 55\%, with details provided in Appendix~\ref{app:ablation}. Using a small sampling number enables faster question selection without substantially compromising knowledge estimate accuracy, which is important for maintaining student engagement in real-world deployment.

\subsection{Qualitative Case Study}
We analyze the predicted student responses conditioned on different KC mastery levels, for the problem \textit{noTeenSum}. Due to spatial limitations, we show the detailed analysis in Appendix~\ref{app:qualititive}. We see that for low KC mastery ($\theta=0$), the GIRT model trained with SFT still generates correct code, differing only in style from the code generated for high mastery ($\theta=1$). We show the generated code in Table~\ref{tab:qual_codeworkout_noTeenSum_sft}.
In contrast, after DPO, the GIRT model generates code containing logical errors when conditioned on low KC mastery. We show the generated code in Table~\ref{tab:qual_codeworkout_noTeenSum_dpo}. These results indicate that DPO is necessary for the GIRT model to effectively leverage student KC mastery levels, which is a prerequisite for the question selection algorithms to differentiate between candidate questions.

\section{Conclusions and Future Work}
In this paper, we proposed CodeGENCAT, a novel Generative CAT framework for coding problems that leverages LLMs to exploit the rich information embedded in programming question text and open-ended student code responses to estimate student knowledge. We showed that our GIRT model can effectively align estimated student knowledge with open-ended student code responses both quantitatively and qualitatively. We also showed that our three novel question-selection algorithms, which utilize the generative capabilities of the GIRT model, outperform existing CAT baselines, particularly in the key early stages of testing.

For the future work, we first plan to investigate our framework on other domains and subjects, such as mathematical reasoning, essay writing, and language learning, depending on data availability. Open-ended questions are common in these domains, which enables us to validate the generalization of our framework. Moreover, we plan to reduce the computation cost of response sampling during the question selection process to facilitate real-time deployment, because minimizing waiting time is necessary to maintain student engagement.  


\clearpage
\newpage

\section*{Limitations} Despite the fact that CodeGENCAT can achieve promising results, CodeGENCAT still has several limitations. First, our experimental evaluation is confined to the domain of introductory computer science education. While CodeGENCAT is theoretically adaptable to other open-ended domains, such as mathematical reasoning or short-answer essays, empirical validation in these areas is currently constrained by the scarcity of publicly available datasets containing student responses to multiple open-ended questions. This scarcity arises primarily from strict privacy regulations (e.g., FERPA, GDPR) protecting sensitive educational records, as well as the high cost of anonymizing unstructured text data, where personal identifiable information is often deeply embedded in student narratives and reasoning. Second, the computational cost of CodeGENCAT is higher than traditional CAT baselines. CodeGENCAT requires generating multiple response samples for every candidate question, which scales linearly with the size of the question bank and may pose latency challenges for real-time deployment. Finally, our current representation of student knowledge relies on expert-annotated KCs, limiting immediate scalability to subjects where such structured metadata is unavailable.

\section*{Ethical Considerations} We acknowledge that integrating LLMs into educational assessment introduces risks regarding bias and privacy. First, the underlying LLM may inherit biases from the specific demographics in the pre-training or fine-tuning datasets, leading to inequitable knowledge estimates across diverse student demographics. Second, analyzing open-ended responses increases privacy risks. While CodeGENCAT utilizes anonymized data, real-world deployment requires strict safeguards to protect sensitive information often embedded in student responses. Therefore, such a system requires rigorous human auditing and extensive field testing prior to real-world deployment. 

We use publicly available pretrained models and datasets in accordance with their corresponding licenses and terms of use. These artifacts are used only for research purposes, consistent with their intended use and access conditions. 

We use two programming education datasets with anonymized student identifiers. We do not use student names, email addresses, or other directly identifying information in our experiments. We also inspect the data preprocessing pipeline to reduce the risk of including personally identifying information or offensive content in released artifacts, and we do not release raw student responses beyond what is permitted by the original dataset access conditions.
\bibliography{main}

\appendix

\section{Scoring Model for Student Response} \label{app:score}
For the scoring model used in Uncertainty question selection algorithm, we use \textbf{Llama-3.2-1B-Instruct} as the backbone LLM. Given a data point $x_{i,j}$ with question text $p_j$ and a student response $c_{i,j}$, our goal is to predict the continuous correctness score $\hat{a}_{i,j} \in [0,1]$ that approximates the ground-truth unit-test outcome $a_{i,j}$.

We formulate the input by concatenating the problem text $p_j$ and student response $c_{i,j}$ into a single prompt sequence. The detailed prompt is in Appendix \ref{tab:scoring_prompt}. To generate a continuous score from the LLM, we adopt the logit-interpolation method described in \cite{scarlatos2025exploring}. Specifically, we extract the logits corresponding to the tokens for ``Correct'' (1) and ``Incorrect'' (0) from LLM's output at the last token position of the input sequence, denoting them as $l_1$ and $l_0$ respectively. We then compute the predicted correctness score $\bar{a}_{i,j}$ by normalizing these two extracted logits using softmax function:
\begin{equation*}
\bar{a}_{i,j} = \frac{e^{l_1}}{e^{l_0} + e^{l_1}}.
\end{equation*}
For a single data point $x_{i,j} \in \mathcal{R}$, we define the loss as the binary cross-entropy between the predicted score $\bar{a}_{i,j}$ and the ground-truth score $a_{i,j}$:
\begin{equation*}
\mathcal{L}_{i,j}^{\text{Score}} = -\bigl[ a_{i,j} \log \bar{a}_{i,j} + (1 - a_{i,j}) \log(1 - \bar{a}_{i,j}) \bigr],
\end{equation*}
Finally, the global training objective for the scoring model is the average loss over all data points:
\begin{equation*}
\mathcal{L}_{\text{Score}} = \frac{1}{|\mathcal{R}|} \sum_{x_{i,j} \in \mathcal{R}} \mathcal{L}_{i,j}^{\text{Score}}.
\end{equation*}
Our trained scoring model achieves a \textbf{Pearson correlation coefficient} \cite{benesty2009pearson} of 0.845 on the CodeWorkout test set and 0.717 on the ProgFeed test set, indicating high agreement with ground-truth scores.

\section{Ablation Study Details} \label{app:ablation}
\paragraph{Time usage for different sampling number}
Although we use vLLM \cite{kwon2025vllm} to accelerate response sampling, reducing $N$ from 10 to 3 still lowers the total adaptive testing time, with the largest reduction observed for Information. For Uncertainty, the runtime decreases from 4:41:11 to 3:41:49, and for Diversity, it decreases from 4:15:54 to 3:33:55. Uncertainty takes slightly longer than Diversity because it evaluates each generated response with a \textbf{Llama-3.2-1B-Instruct} scoring model, whereas Diversity only computes pairwise similarity using the smaller pre-trained \textbf{CodeBERT} embedding model. Information has the highest runtime reduction, decreasing from 15:20:35 with $N=10$ to 6:51:28 with $N=3$, because it computes Fisher information by passing each sampled response through the full GIRT model.
 Therefore, reducing $N$ from 10 to 3 lowers inference time by up to approximately 55\%, enabling faster question selection without substantially compromising knowledge estimate accuracy.

\begin{table*}[t]
\centering
\resizebox{\textwidth}{!}{
\begin{tabular}{l ccccccc c ccccccc}
\toprule
\multirow{3}{*}{\textbf{Method}} & \multicolumn{15}{c}{\textbf{Step $t$}} \\
\cmidrule(lr){2-16}
& \multicolumn{7}{c}{\textbf{CodeWorkout}} && \multicolumn{7}{c}{\textbf{ProgFeed}} \\
\cmidrule(lr){2-8} \cmidrule(lr){10-16}
& $1$ & $3$ & $5$ & $7$ & $10$ & $15$ & $20$ && $1$ & $3$ & $5$ & $7$ & $10$ & $15$ & $20$ \\
\midrule
\multicolumn{16}{c}{\textbf{AUC $\uparrow$}} \\
\midrule
1B Uncertainty & \textbf{0.6737} & \textbf{0.7258} & 0.7415 & 0.7599 & 0.7693 & 0.7838 & 0.7873 && \textbf{0.6375} & \textbf{0.6725} & \textbf{0.6875} & \textbf{0.6930} & 0.7020 & 0.7063 & 0.7101 \\
3B Uncertainty & 0.6612 & 0.7121 & \textbf{0.7531} & \textbf{0.7626} & \textbf{0.7746} & \textbf{0.7845} & \textbf{0.7893} && 0.6091 & 0.6582 & 0.6789 & 0.6852 & \textbf{0.7081} & \textbf{0.7175} & \textbf{0.7200} \\
\cmidrule(lr){1-16}
1B Diversity   & 0.7073 & 0.7352 & 0.7529 & 0.7632 & 0.7687 & 0.7801 & 0.7855 && 0.6437 & 0.6319 & 0.6834 & 0.6870 & 0.6901 & \textbf{0.7064} & \textbf{0.7161} \\
3B Diversity   & \textbf{0.7235} & \textbf{0.7355} & \textbf{0.7547} & \textbf{0.7673} & \textbf{0.7777} & \textbf{0.7833} & \textbf{0.7910} && \textbf{0.6493} & \textbf{0.6601} & \textbf{0.6918} & \textbf{0.6924} & \textbf{0.7022} & 0.7029 & 0.7131 \\
\cmidrule(lr){1-16}
1B Information & \textbf{0.6854} & 0.7264 & 0.7492 & 0.7616 & 0.7718 & 0.7842 & \textbf{0.7901} && \textbf{0.6730} & \textbf{0.6845} & \textbf{0.6910} & \textbf{0.7080} & 0.7056 & 0.7204 & \textbf{0.7191} \\
3B Information & 0.6851 & \textbf{0.7476} & \textbf{0.7655} & \textbf{0.7717} & \textbf{0.7787} & \textbf{0.7861} & 0.7898 && 0.6400 & 0.6551 & 0.6903 & 0.6971 & \textbf{0.7069} & \textbf{0.7206} & 0.7149 \\
\midrule
\multicolumn{16}{c}{\textbf{Accuracy $\uparrow$}} \\
\midrule
1B Uncertainty & 0.6012 & \textbf{0.6796} & 0.6788 & \textbf{0.6980} & 0.6988 & 0.7148 & 0.7212 && \textbf{0.6296} & 0.7248 & 0.7448 & 0.7432 & 0.7616 & 0.7560 & 0.7656 \\
3B Uncertainty & \textbf{0.6020} & 0.6636 & \textbf{0.6944} & \textbf{0.6980} & \textbf{0.7136} & \textbf{0.7180} & \textbf{0.7232} && 0.5928 & \textbf{0.7288} & \textbf{0.7464} & \textbf{0.7696} & \textbf{0.7680} & \textbf{0.7632} & \textbf{0.7696} \\
\cmidrule(lr){1-16}
1B Diversity   & 0.6452 & 0.6720 & 0.6888 & 0.6892 & 0.6984 & 0.7020 & 0.7136 && 0.6856 & 0.7176 & 0.7424 & 0.7496 & 0.7600 & \textbf{0.7704} & \textbf{0.7728} \\
3B Diversity   & \textbf{0.6532} & \textbf{0.6736} & \textbf{0.6940} & \textbf{0.7012} & \textbf{0.7124} & \textbf{0.7168} & \textbf{0.7220} && \textbf{0.7224} & \textbf{0.7376} & \textbf{0.7608} & \textbf{0.7616} & \textbf{0.7688} & \textbf{0.7704} & 0.7680 \\
\cmidrule(lr){1-16}
1B Information & \textbf{0.6328} & 0.6740 & 0.6876 & 0.6964 & 0.7044 & 0.7068 & \textbf{0.7232} && \textbf{0.6960} & 0.7184 & 0.7456 & \textbf{0.7648} & \textbf{0.7648} & \textbf{0.7760} & \textbf{0.7728} \\
3B Information & 0.6160 & \textbf{0.6892} & \textbf{0.6968} & \textbf{0.7012} & \textbf{0.7076} & \textbf{0.7176} & 0.7212 && 0.6792 & \textbf{0.7448} & \textbf{0.7664} & 0.7640 & 0.7584 & 0.7720 & 0.7696 \\
\bottomrule
\end{tabular}
}
\caption{5-fold averaged CodeGENCAT results under different GIRT backbone sizes (1B vs.\ 3B) across question selection algorithms. Bold indicates the better result between 1B and 3B for the same question selection algorithm at each step.}
\label{tab:gencat_1b_3b_ablation}
\end{table*}

\begin{table}[t]
\centering
\scriptsize
\setlength{\tabcolsep}{2.5pt}
\scalebox{.97}{
\begin{tabular}{l ccccccc}
\toprule
\multirow{2}{*}{\textbf{Method}} & \multicolumn{7}{c}{\textbf{CodeWorkout}} \\
\cmidrule(lr){2-8}
& $1$ & $3$ & $5$ & $7$ & $10$ & $15$ & $20$ \\
\midrule
\multicolumn{8}{c}{\textbf{AUC $\uparrow$}} \\
\midrule
DPO Uncertainty & 0.6612 & 0.7121 & \textbf{0.7531} & 0.7626 & 0.7746 & 0.7845 & \textbf{0.7893} \\
SFT Uncertainty & \textbf{0.6768} & \textbf{0.7146} & 0.7389 & \textbf{0.7632} & \textbf{0.7749} & \textbf{0.7870} & \textbf{0.7893} \\
\cmidrule(lr){1-8}
DPO Diversity   & \textbf{0.7235} & \textbf{0.7355} & \textbf{0.7547} & \textbf{0.7673} & \textbf{0.7777} & \textbf{0.7833} & \textbf{0.7910} \\
SFT Diversity   & 0.7079 & 0.7277 & 0.7444 & 0.7563 & 0.7691 & 0.7821 & 0.7872 \\
\cmidrule(lr){1-8}
DPO Information & 0.6851 & \textbf{0.7476} & \textbf{0.7655} & \textbf{0.7717} & 0.7787 & \textbf{0.7861} & 0.7898 \\
SFT Information & \textbf{0.6911} & 0.7380 & 0.7616 & 0.7713 & \textbf{0.7813} & 0.7843 & \textbf{0.7914} \\
\midrule
\multicolumn{8}{c}{\textbf{Accuracy $\uparrow$}} \\
\midrule
DPO Uncertainty & 0.6020 & 0.6636 & \textbf{0.6944} & 0.6980 & \textbf{0.7136} & 0.7180 & \textbf{0.7232} \\
SFT Uncertainty & \textbf{0.6124} & \textbf{0.6656} & 0.6784 & \textbf{0.7032} & 0.7072 & \textbf{0.7280} & 0.7224 \\
\cmidrule(lr){1-8}
DPO Diversity   & \textbf{0.6532} & \textbf{0.6736} & \textbf{0.6940} & 0.7012 & \textbf{0.7124} & 0.7168 & \textbf{0.7220} \\
SFT Diversity   & 0.6432 & 0.6732 & 0.6844 & \textbf{0.7020} & 0.7012 & \textbf{0.7196} & 0.7192 \\
\cmidrule(lr){1-8}
DPO Information & 0.6160 & \textbf{0.6892} & \textbf{0.6968} & 0.7012 & \textbf{0.7076} & \textbf{0.7176} & 0.7212 \\
SFT Information & \textbf{0.6192} & 0.6824 & 0.6884 & \textbf{0.7076} & 0.7072 & 0.7140 & \textbf{0.7216} \\
\bottomrule
\end{tabular}}
\caption{5-fold averaged CodeGENCAT results under different GIRT training stages (SFT vs.\ DPO) across question selection algorithms. Bold indicates the better result between SFT and DPO for the same question selection algorithm at each step.}
\label{tab:codeworkout_3b_dpo_sft_ablation}
\end{table}

\begin{table}[t]
\centering
\scriptsize
\setlength{\tabcolsep}{2.5pt}
\scalebox{.9}{
\begin{tabular}{l ccccccc}
\toprule
\multirow{2}{*}{\textbf{Method}} & \multicolumn{7}{c}{\textbf{CodeWorkout }} \\
\cmidrule(lr){2-8}
& $1$ & $3$ & $5$ & $7$ & $10$ & $15$ & $20$ \\
\midrule
\multicolumn{8}{c}{\textbf{AUC $\uparrow$}} \\
\midrule
Sample 3 Uncertainty  & \textbf{0.6918} & \textbf{0.7329} & 0.7522 & 0.7616 & 0.7734 & 0.7772 & 0.7834 \\
Sample 10 Uncertainty & 0.6612 & 0.7121 & \textbf{0.7531} & \textbf{0.7626} & \textbf{0.7746} & \textbf{0.7845} & \textbf{0.7893} \\
\cmidrule(lr){1-8}
Sample 3 Diversity    & 0.6927 & \textbf{0.7424} & \textbf{0.7645} & \textbf{0.7703} & 0.7746 & \textbf{0.7856} & 0.7876 \\
Sample 10 Diversity   & \textbf{0.7235} & 0.7355 & 0.7547 & 0.7673 & \textbf{0.7777} & 0.7833 & \textbf{0.7910} \\
\cmidrule(lr){1-8}
Sample 3 Information  & \textbf{0.6927} & 0.7421 & 0.7612 & 0.7652 & 0.7766 & \textbf{0.7877} & \textbf{0.7908} \\
Sample 10 Information & 0.6851 & \textbf{0.7476} & \textbf{0.7655} & \textbf{0.7717} & \textbf{0.7787} & 0.7861 & 0.7898 \\
\midrule
\multicolumn{8}{c}{\textbf{Accuracy $\uparrow$}} \\
\midrule
Sample 3 Uncertainty  & \textbf{0.6284} & \textbf{0.6800} & 0.6856 & \textbf{0.7016} & 0.7124 & 0.7148 & 0.7216 \\
Sample 10 Uncertainty & 0.6020 & 0.6636 & \textbf{0.6944} & 0.6980 & \textbf{0.7136} & \textbf{0.7180} & \textbf{0.7232} \\
\cmidrule(lr){1-8}
Sample 3 Diversity    & 0.6264 & \textbf{0.6952} & \textbf{0.6984} & \textbf{0.7068} & 0.7084 & 0.7136 & 0.7172 \\
Sample 10 Diversity   & \textbf{0.6532} & 0.6736 & 0.6940 & 0.7012 & \textbf{0.7124} & \textbf{0.7168} & \textbf{0.7220} \\
\cmidrule(lr){1-8}
Sample 3 Information  & \textbf{0.6232} & \textbf{0.6916} & \textbf{0.7004} & \textbf{0.7056} & 0.7060 & 0.7140 & 0.7132 \\
Sample 10 Information & 0.6160 & 0.6892 & 0.6968 & 0.7012 & \textbf{0.7076} & \textbf{0.7176} & \textbf{0.7212} \\
\bottomrule
\end{tabular}}
\caption{5-fold averaged CodeGENCAT results under different sampling numbers (3 vs.\ 10) across question selection algorithms. Bold indicates the better result between \_3 and \_10 for the same question selection algorithm at each step.}
\label{tab:codeworkout_sample_ablation}
\end{table}

\section{Knowledge Estimation Accuracy with Response Similarity} \label{app:codemetric}

\begin{table*}[t]
\centering
\scriptsize
\setlength{\tabcolsep}{3pt}
\resizebox{\textwidth}{!}{
\begin{tabular}{ll ccccccc c ccccccc}
\toprule
\multirow{4}{*}{\textbf{Model}} & \multirow{4}{*}{\textbf{Method}} 
& \multicolumn{15}{c}{\textbf{Step $t$}} \\
\cmidrule(lr){3-17}
& & \multicolumn{15}{c}{\textbf{CodeBLEU $\uparrow$}} \\
\cmidrule(lr){3-17}
& & \multicolumn{7}{c}{\textbf{CodeWorkout}} && \multicolumn{7}{c}{\textbf{ProgFeed}} \\
\cmidrule(lr){3-9} \cmidrule(lr){11-17}
& & $1$ & $3$ & $5$ & $7$ & $10$ & $15$ & $20$ && $1$ & $3$ & $5$ & $7$ & $10$ & $15$ & $20$ \\
\midrule
\multirow{7}{*}{3B}
& Uncertainty & 0.5598 & \textbf{0.5807} & 0.5845 & 0.5827 & 0.5902 & 0.5899 & 0.5915 && 0.4852 & \textbf{0.4865} & \textbf{0.4855} & 0.4855 & 0.4850 & 0.4850 & 0.4846 \\
& Diversity   & 0.5638 & 0.5797 & 0.5845 & 0.5887 & 0.5898 & 0.5905 & 0.5901 && 0.4855 & 0.4861 & 0.4852 & 0.4855 & \textbf{0.4874} & 0.4847 & 0.4848 \\
& Information & \textbf{0.5669} & 0.5775 & 0.5855 & 0.5901 & 0.5906 & 0.5917 & \textbf{0.5947} && \textbf{0.4860} & 0.4861 & 0.4853 & \textbf{0.4863} & 0.4859 & \textbf{0.4858} & \textbf{0.4860} \\
& 1PL IRT     & 0.5581 & 0.5776 & 0.5797 & 0.5876 & \textbf{0.5914} & 0.5905 & 0.5923 && 0.4851 & 0.4835 & 0.4841 & 0.4852 & 0.4836 & 0.4842 & 0.4839 \\
& NCAT        & 0.5638 & 0.5798 & \textbf{0.5872} & \textbf{0.5916} & 0.5910 & \textbf{0.5941} & 0.5939 && 0.4835 & 0.4836 & 0.4847 & 0.4829 & 0.4853 & 0.4835 & 0.4848 \\
& LACAT       & 0.5319 & 0.5608 & 0.5681 & 0.5772 & 0.5778 & 0.5813 & 0.5804 && 0.4833 & 0.4835 & 0.4844 & 0.4838 & 0.4853 & 0.4838 & 0.4852 \\
& BOBCAT      & 0.5307 & 0.5594 & 0.5665 & 0.5766 & 0.5865 & 0.5904 & 0.5935 && 0.4844 & 0.4841 & 0.4851 & 0.4855 & 0.4856 & 0.4841 & 0.4850 \\
\cmidrule(lr){1-17}
\multirow{7}{*}{1B}
& Uncertainty & 0.5528 & 0.5663 & 0.5817 & 0.5899 & 0.5909 & 0.5913 & 0.5913 && 0.2194 & 0.2202 & 0.2203 & \textbf{0.2210} & 0.2206 & 0.2207 & 0.2206 \\
& Diversity   & 0.5516 & 0.5704 & 0.5868 & 0.5872 & 0.5891 & 0.5889 & 0.5892 && 0.2196 & 0.2207 & 0.2185 & 0.2185 & 0.2189 & 0.2187 & 0.2184 \\
& Information & \textbf{0.5566} & 0.5691 & 0.5861 & 0.5905 & 0.5927 & \textbf{0.5973} & \textbf{0.5991} && 0.2197 & 0.2198 & 0.2197 & 0.2196 & 0.2194 & 0.2207 & 0.2208 \\
& 1PL IRT     & 0.5535 & 0.5670 & 0.5844 & 0.5870 & 0.5906 & 0.5902 & 0.5919 && 0.2202 & 0.2212 & 0.2197 & 0.2208 & 0.2201 & 0.2207 & 0.2196 \\
& NCAT        & 0.5513 & \textbf{0.5707} & \textbf{0.5895} & \textbf{0.5938} & \textbf{0.5936} & 0.5946 & 0.5929 && 0.2210 & 0.2202 & 0.2204 & 0.2204 & \textbf{0.2210} & 0.2200 & 0.2203 \\
& LACAT       & 0.5490 & 0.5637 & 0.5816 & 0.5869 & 0.5862 & 0.5910 & 0.5910 && \textbf{0.2211} & \textbf{0.2219} & \textbf{0.2216} & 0.2204 & 0.2206 & \textbf{0.2211} & \textbf{0.2210} \\
& BOBCAT      & 0.5485 & 0.5611 & 0.5726 & 0.5787 & 0.5844 & 0.5899 & 0.5937 && 0.2193 & 0.2210 & 0.2202 & 0.2204 & 0.2195 & 0.2201 & 0.2202 \\
\bottomrule
\end{tabular}
}
\caption{5-fold averaged Code similarity results (CodeBLEU) on CodeWorkout and ProgFeed across varying test lengths ($t$) under the 1B and 3B GIRT models and baselines. Best results are in \textbf{bold}.}
\label{tab:codebleu_1b_3b_both_datasets}
\end{table*}

\begin{table*}[t]
\centering
\scriptsize
\setlength{\tabcolsep}{3pt}
\resizebox{\textwidth}{!}{
\begin{tabular}{ll ccccccc c ccccccc}
\toprule
\multirow{4}{*}{\textbf{Model}} & \multirow{4}{*}{\textbf{Method}} 
& \multicolumn{15}{c}{\textbf{Step $t$}} \\
\cmidrule(lr){3-17}
& & \multicolumn{15}{c}{\textbf{Pearson Correlation $\uparrow$}} \\
\cmidrule(lr){3-17}
& & \multicolumn{7}{c}{\textbf{CodeWorkout}} && \multicolumn{7}{c}{\textbf{ProgFeed}} \\
\cmidrule(lr){3-9} \cmidrule(lr){11-17}
& & $1$ & $3$ & $5$ & $7$ & $10$ & $15$ & $20$ && $1$ & $3$ & $5$ & $7$ & $10$ & $15$ & $20$ \\
\midrule
\multirow{7}{*}{3B}
& Uncertainty & 0.1909 & 0.2518 & 0.2635 & 0.2640 & 0.2758 & 0.2862 & 0.2821 && 0.1353 & 0.1454 & 0.1502 & 0.1231 & 0.1246 & 0.1301 & 0.1307 \\
& Diversity   & \textbf{0.2117} & \textbf{0.2721} & 0.2747 & 0.2769 & 0.2839 & 0.2871 & 0.2815 && 0.1348 & 0.1493 & \textbf{0.1597} & 0.1359 & 0.1296 & 0.1318 & 0.1390 \\
& Information & 0.1799 & 0.2714 & \textbf{0.2831} & \textbf{0.3095} & \textbf{0.2996} & \textbf{0.3029} & \textbf{0.3206} && \textbf{0.1439} & 0.1621 & 0.1266 & 0.1403 & 0.1297 & \textbf{0.1549} & 0.1334 \\
& 1PL IRT     & 0.1764 & 0.2591 & 0.2471 & 0.2687 & 0.2854 & 0.2817 & 0.2913 && 0.1359 & \textbf{0.1773} & 0.1526 & 0.1244 & 0.1387 & 0.1295 & 0.1309 \\
& NCAT        & 0.1573 & 0.2396 & 0.2469 & 0.2814 & 0.2725 & 0.2771 & 0.2836 && 0.1378 & 0.1389 & 0.1374 & \textbf{0.1455} & \textbf{0.1400} & 0.1384 & \textbf{0.1411} \\
& LACAT       & 0.1470 & 0.2049 & 0.2269 & 0.2376 & 0.2565 & 0.2500 & 0.2644 && 0.1342 & 0.1286 & 0.1289 & 0.1294 & 0.1263 & 0.1340 & 0.1313 \\
& BOBCAT      & 0.1564 & 0.1956 & 0.2141 & 0.2517 & 0.2439 & 0.2795 & 0.2859 && 0.1335 & 0.1318 & 0.1271 & 0.1276 & 0.1276 & 0.1386 & 0.1373 \\
\cmidrule(lr){1-17}
\multirow{7}{*}{1B}
& Uncertainty & 0.1098 & 0.1462 & 0.1945 & 0.2189 & 0.2183 & 0.2353 & 0.2334 && 0.1162 & \textbf{0.1331} & 0.1160 & 0.1027 & 0.1077 & 0.1192 & 0.1204 \\
& Diversity   & 0.1106 & \textbf{0.1825} & 0.2181 & 0.2378 & 0.2368 & 0.2399 & 0.2393 && \textbf{0.1178} & 0.1054 & 0.1094 & 0.1089 & 0.0991 & 0.1044 & 0.1148 \\
& Information & \textbf{0.1194} & 0.1746 & \textbf{0.2243} & \textbf{0.2423} & \textbf{0.2536} & \textbf{0.2557} & \textbf{0.2508} && 0.1115 & 0.1154 & 0.1186 & 0.1192 & 0.1157 & 0.1023 & 0.1054 \\
& 1PL IRT     & 0.1111 & 0.1693 & 0.2066 & 0.2152 & 0.2308 & 0.2346 & 0.2421 && 0.1083 & 0.1018 & 0.1172 & 0.1101 & 0.0907 & 0.1073 & 0.1175 \\
& NCAT        & 0.1039 & 0.1628 & 0.2103 & 0.2235 & 0.2359 & 0.2400 & 0.2449 && 0.1119 & 0.1021 & 0.1127 & 0.1042 & 0.1143 & 0.1087 & 0.1115 \\
& LACAT       & 0.1040 & 0.1426 & 0.1829 & 0.2221 & 0.2234 & 0.2269 & 0.2373 && 0.1119 & 0.1167 & 0.1221 & 0.1100 & 0.1086 & 0.0900 & 0.0849 \\
& BOBCAT      & 0.1078 & 0.1325 & 0.1620 & 0.1906 & 0.2082 & 0.2409 & 0.2462 && 0.1120 & 0.1184 & \textbf{0.1313} & \textbf{0.1308} & \textbf{0.1200} & \textbf{0.1209} & \textbf{0.1294} \\
\bottomrule
\end{tabular}
}
\caption{5-fold averaged Pearson correlation coefficient results on CodeWorkout and ProgFeed across varying test lengths ($t$) under the 1B and 3B GIRT models and baselines. Best results are in \textbf{bold}.}
\label{tab:pearson_1b_3b_both_datasets}
\end{table*}

Besides evaluating the accuracy of the student knowledge estimate via binary classification, we also test the knowledge estimate accuracy by checking the similarity of the generated student responses based on the knowledge estimate to the ground truth responses on the test set. For each framework, we first use the questions selected by that framework to update the student knowledge estimate. We then use the GIRT model to generate response based on the updated knowledge estimate. Thus, for both CodeGENCAT and the baselines, the difference is the selected questions. To capture both the structural and functional aspects of the generated response, we evaluate these responses from two perspectives. The first perspective utilizes \textbf{CodeBLEU} \cite{ren2020codebleu} to measure structural similarity by combining traditional n-gram BLEU matching with code-specific syntactic and semantic features, such as abstract syntax tree matching. The second perspective evaluates functional correctness using the \textbf{Pearson correlation coefficient}. Specifically, we utilize our trained scoring model to assign a score to the predicted response and calculate the correlation between this predicted score and the ground truth score.

We report the 5-fold averaged CodeBLEU score of CodeGENCAT and four baselines across test lengths ranging from $t=1$ to $t=20$, on both 1B and 3B GIRT models and two datasets, in Table \ref{tab:codebleu_1b_3b_both_datasets}. On the CodeWorkout dataset, we observe that code generation similarity consistently improves as the test length $t$ increases. For instance, the CodeBLEU score for GENCAT (Information) with 1B GIRT model rises from 0.5566 at $t=1$ to 0.5927 at $t=10$. This result shows that with more questions getting selected, the student knowledge is estimated more accurately, resulting in generating responses that is more similar to the ground truth responses. However, the performance gaps between different question selection algorithms at the same time step remain relatively small. We think this small gap is due to the high structural overlap between the generated responses of different question selection algorithms, as illustrated in our qualitative analysis in Appendix \ref{tab:qual_codeworkout_noTeenSum_dpo}. In the \textit{noTeenSum} problem, the GIRT model generates a specific logic error for a student with explicit knowledge = 0, and the correct code for a student with explicit knowledge = 0.5. Despite the opposite correctness outcomes, these two code snippets share identical method signatures, variable names, and code structures, differing by only a few lines of logic. Consequently, CodeBLEU struggles to distinguish between these semantically different but structurally similar responses, resulting in similar scores across different question selection algorithms. 

On the ProgFeed dataset, the differences in CodeBLEU scores across time steps is minimal. A potential reason is that the programming questions in ProgFeed are simpler and more straightforward than those in CodeWorkout. This simplicity leads to minimal room for variation in terms of how the code can be written, both semantically and structurally. Finally, when comparing the GIRT model size, on the Progfeed dataset, the 3B model achieves higher CodeBLEU scores than the 1B model. On the CodeWorkout dataset, the scores are similar. This result indicates that the increased capacity of the 3B model allows it to generate code that aligns more accurately with the structural and semantic patterns of the ground truth responses.


We also report the 5-fold averaged Pearson correlation coefficient in Table \ref{tab:pearson_1b_3b_both_datasets}. For the 3B GIRT model on the CodeWorkout dataset, CodeGENCAT has a clear performance advantage over the baselines. For example, at $t=5$, Diversity outperforms the strongest baseline, 1PL IRT, by a relatively large margin (0.2831 vs. 0.2471). This result indicates that the responses generated by CodeGENCAT are more aligned with the actual functional behavior of ground truth responses. However, this improvement becomes smaller when using the smaller 1B GIRT model. At $t=5$, Diversity outperforms NCAT, but the gap is smaller (0.2243 vs. 0.2103). This result suggests that CodeGENCAT benefits more from a larger and more capable GIRT model. Furthermore, on the ProgFeed dataset, the differences between CodeGENCAT and the baselines are minimal across both model sizes. These findings suggest that while using a more capable model on student responses to more difficult questions is helpful, smaller models or simpler questions limit CodeGENCAT's ability to show significant improvement over existing CAT methods.

\section{Additional Experimental Details} \label{app:exp}
We perform a preliminary hyperparameter search on our validation sets for different CodeGENCAT components. We use ChatGPT \cite{achiam2023gpt} to help us select potential hyperparameter values.
\subsection{GIRT model Training Details} 
\subsubsection{SFT}
On CodeWorkout dataset, the GIRT model is fine-tuned for 15 epochs with a batch size of 4 and 16 gradient accumulation steps. On Progfeed dataset, the GIRT model is fine-tuned for 3 epochs with a batch size of 4 and 16 gradient accumulation steps. The Llama backbone LLM is trained with a learning rate of $1 \times 10^{-5}$, while the latent knowledge parameters and MLP are trained with a learning rate of $0.01$. we utilize the AdamW optimizer for all parameters. A linear warmup is applied for the first 3\% of training steps. We set the dimension of latent knowledge $\mathbf{z}_i \in \mathbb{R}^D$ where $D= 2$, and we set $\lambda = 0.2$ to balance the two training losses. Training requires approximately 1 hour and 23 minutes on one NVIDIA L40 GPU on CodeWorkout and 24 minutes on Progfeed.  
\subsubsection{DPO}
On both datasets, the GIRT model is trained with DPO objective for 1 epoch with a batch size of 2 and 16 gradient accumulation steps. We utilize the AdamW optimizer with learning rate = $1 \times 10^{-6}$. We set the KL-divergence penalty parameter $\beta = 0.5$. For constructing preference pairs, we set $M=3$ and $\tau=0.1$ to select rejected samples. Training requires approximately 25 minutes on one NVIDIA L40 GPU. 

\subsection{Adaptive Testing Cycle Details}
During the adaptive testing cycle, we set the number of sampled responses $N=10$ for each candidate question. The responses are generated using nucleus sampling with a temperature of 0.8, top-$p$ of 0.9, and top-$k$ of 50. To update the student knowledge estimate $\mathbf{z}_i$, we utilize the AdamW optimizer with a learning rate of $0.01$. We utilize an incremental knowledge update strategy where the number of optimization epochs scales with the test sequence length (2 epochs per step), capped at a maximum of 10 epochs. Consistent with the SFT objective, we maintain the loss balancing coefficient $\lambda = 0.2$ during the knowledge update phase. Adaptive testing time varies based on the test sequence length on one NVIDIA L40 GPU.

\section{Data Splitting Details} \label{app:data}
For both GIRT model training and CAT evaluation group, we implement a specific data splitting that ensures KC coverage. We employ a greedy set cover algorithm to partition each student responses. This algorithm first identifies the minimal set of questions required to cover all KCs associated with the student responses and assigns them to the training set (for the GIRT model training group) or the candidate question set (for the CAT evaluation group). This constraint guarantees that every KC appearing in the evaluation or held-out test set is also present in the training or candidate sets. This constraint can prevent the cold-start problem for specific KCs, ensuring the student mastery level on a KC is estimated before being tested.

We then apply group-specific splitting rules. For the GIRT model training group, after the greedy set cover assigns the necessary questions to the training set, the remaining questions are shuffled and allocated according to the dataset-specific split. For CodeWorkout, we allocate the remaining responses to obtain an approximate 80:10:10 train/validation/test split. For ProgFeed, we use a 70:30 train/test split.

For the CAT evaluation group, we determine the test set size by calculating 20\% of the average number of responses across students, yielding 10 held-out test questions on CodeWorkout and 5 on ProgFeed. The remaining questions form the candidate question set for adaptive selection. We require each CAT evaluation student to have at least 30 candidate questions on CodeWorkout and at least 20 candidate questions on ProgFeed; otherwise, the student is reassigned to the GIRT model training group. 

\section{GIRT model Qualitative Analysis} \label{app:qualititive}
In this section, we qualitatively analyze the sensitivity of our GIRT model. For CodeGENCAT to function effectively, the GIRT model must be sensitive regarding student estimated KC mastery levels: it should generate incorrect responses when conditioned on low KC mastery levels ($\theta \approx 0$) and correct responses when conditioned on high mastery levels ($\theta \approx 1$). If the model generates the same responses regardless of the KC mastery levels, the question-selection algorithms cannot differentiate between candidate questions. We analyze the model's response for a problem under three mastery levels ($\theta \in \{0, 0.5, 1\}$), comparing the SFT baseline against our DPO-refined model. 
Table~\ref{tab:qual_codeworkout_noTeenSum_sft} in the Appendix demonstrates that the SFT model fails to condition generation on student KC mastery levels. Even when the input KC mastery levels are set to 0, the model generates the correct code. While the code structure differs slightly across KC mastery levels, the model consistently outputs correct code.
In contrast, Table~\ref{tab:qual_codeworkout_noTeenSum_dpo} demonstrates that the DPO-refined model can conditions generation on student KC mastery levels. When the input KC mastery levels are set to 0, the model generates code with a logic error: it incorrectly returns 0 for number 15 and 16. When conditioned on higher KC mastery levels, the model generates correct code, while the coding style differs between mastery levels 0.5 and 1. This difference shows that DPO is necessary to enforce the GIRT model to conditions generation on student KC mastery levels, which is a fundamental requirement for the CodeGENCAT question-selection algorithms to work.

\section{Package Implementations and Parameters.}
For the packages used in data preprocessing, model training, and evaluation, we use publicly available implementations from their official or commonly used repositories. For CodeBLEU evaluation, we use the provided implementation with the programming language set to Java for CodeWorkout and Python for ProgFeed, and use component weights of $0.25$ for n-gram match, weighted n-gram match, syntax match, and data-flow match. We will release our code and configuration files to facilitate reproducibility.
\clearpage
\onecolumn
\section{Qualitative Analysis Example} 
\noindent\rule{\textwidth}{0.8pt}

\noindent\textbf{Question:}
Write two methods in Java that implements the following logic: Given 3 int values, $a$, $b$, and $c$, return their sum.
However, if any of the values is a teen--in the range 13..19 inclusive--then that value counts as 0, except 15 and 16 do not count as teens.
Write a separate helper method called \texttt{fixTeen()} that takes in an int value and returns that value fixed for the teen rule.

\noindent\rule{\textwidth}{0.8pt}

\noindent\textbf{Generative Item Response Model after DPO}

\setlength{\tabcolsep}{8pt}
\renewcommand{\arraystretch}{1.25}

\begin{longtable}{>{\centering\arraybackslash}p{4.2cm} >{\raggedright\arraybackslash}p{11.0cm}}
\toprule
\textbf{Student KC mastery} & \textbf{Code} \\
\midrule
\endfirsthead

\toprule
\textbf{Student KC mastery} & \textbf{Code} \\
\midrule
\endhead

\bottomrule
\endfoot

\multirow{1}{*}{\textbf{0}} &
\begin{minipage}[t]{\linewidth}\vspace{2pt}
\begin{verbatim}
public int noTeenSum(int a, int b, int c)
{
    a = fixTeen(a);
    b = fixTeen(b);
    c = fixTeen(c);
    return a + b + c;
}

public int fixTeen(int n)
{
    if (n >= 13 && n <= 19)
    {
        if (n == 15)
        {
            n = 0;
        }
        else
        {
            n = 0;
        }
    }
    else
    {
        n = n;
    }
    return n;
}
\end{verbatim}
\end{minipage}
\\

\multirow{1}{*}{\textbf{0.5}} &
\begin{minipage}[t]{\linewidth}\vspace{2pt}
\begin{verbatim}
public int noTeenSum(int a, int b, int c)
{
    a = fixTeen(a);
    b = fixTeen(b);
    c = fixTeen(c);
    return a + b + c;
}

public int fixTeen(int n)
{
    if (n >= 13 && n <= 19)
    {
        if (n == 15 || n == 16)
        {
            return n;
        }
        else
        {
            return 0;
        }
    }
    else
    {
        return n;
    }
}
\end{verbatim}
\end{minipage}
\\

\multirow{1}{*}{\textbf{1}} &
\begin{minipage}[t]{\linewidth}\vspace{2pt}
\begin{verbatim}
public int noTeenSum(int a, int b, int c)
{
    return (fixTeen(a) + fixTeen(b) + fixTeen(c));
}

public int fixTeen(int n)
{
    if(n >= 13 && n <= 19)
    {
        if(n == 15 || n == 16)
        {
            return n;
        }
        else
        {
            return 0;
        }
    }
    else
    {
        return n;
    }
}
\end{verbatim}
\end{minipage}
\\

\bottomrule
\caption{Generated code for problem \texttt{noTeenSum} using the DPO-refined model across varying KC mastery levels.}
\vspace{8pt}
\label{tab:qual_codeworkout_noTeenSum_dpo}
\end{longtable}

\clearpage

\noindent\rule{\textwidth}{0.8pt}
\noindent\textbf{Generative Item Response Model after SFT}

\setlength{\tabcolsep}{8pt}
\renewcommand{\arraystretch}{1.25}

\begin{longtable}{>{\centering\arraybackslash}p{4.2cm} >{\raggedright\arraybackslash}p{11.0cm}}
\toprule
\textbf{Student KC mastery} & \textbf{Code} \\
\midrule
\endfirsthead

\toprule
\textbf{Student KC mastery} & \textbf{Code} \\
\midrule
\endhead

\bottomrule
\endfoot

\multirow{1}{*}{\textbf{0}} &
\begin{minipage}[t]{\linewidth}\vspace{2pt}
\begin{verbatim}
public int noTeenSum(int a, int b, int c)
{
    a = fixTeen(a);
    b = fixTeen(b);
    c = fixTeen(c);
    return a + b + c;
}

public int fixTeen(int n)
{
    if (n >= 13 && n <= 19)
    {
        if (n == 15)
        {
            return 15;
        }
        else if (n == 16)
        {
            return 16;
        }
        else
        {
            return 0;
        }
    }
    else
    {
        return n;
    }
}
\end{verbatim}
\end{minipage}
\\

\multirow{1}{*}{\textbf{0.5}} &
\begin{minipage}[t]{\linewidth}\vspace{2pt}
\begin{verbatim}
public int noTeenSum(int a, int b, int c)
{
    a = fixTeen(a);
    b = fixTeen(b);
    c = fixTeen(c);
    return a + b + c;
}

public int fixTeen(int n)
{
    if (n >= 13 && n <= 19)
    {
        if (n == 15 || n == 16)
        {
            return n;
        }
        else
        {
            return 0;
        }
    }
    else
    {
        return n;
    }
}
\end{verbatim}
\end{minipage}
\\

\multirow{1}{*}{\textbf{1}} &
\begin{minipage}[t]{\linewidth}\vspace{2pt}
\begin{verbatim}
public int noTeenSum(int a, int b, int c)
{
    a = fixTeen(a);
    b = fixTeen(b);
    c = fixTeen(c);
    return a + b + c;
}

public int fixTeen(int n)
{
    if (n >= 13 && n <= 19)
    {
        if (n == 15 || n == 16)
        {
            return n;
        }
        else
        {
            return 0;
        }
    }
    else
    {
        return n;
    }
}
\end{verbatim}
\end{minipage}
\\

\bottomrule
\caption{Generated code for problem \texttt{noTeenSum} using the SFT baseline model across varying KC mastery levels.}
\vspace{8pt}
\label{tab:qual_codeworkout_noTeenSum_sft}
\end{longtable}

\clearpage
\onecolumn
\section{Prompt Format}
We provide the prompts for GIRT model and scoring model. We use $<>$ to indicate that a variable is filled in dynamically.

\begin{longtable}{p{15.0cm}}
\toprule
System:\newline
You are a helpful assistant.\newline
You generate plausible student-written Java code given a coding problem and the student knowledge of knowledge components (KCs) that are related to the problem.\newline
Correctness is not required — the code should reflect the student partial understanding, and may contain inefficiencies, bugs, or non-idiomatic style.\newline\newline
User:\newline
Given a Java coding problem and the student knowledge on several knowledge components (KCs) that are related to the problem, your job is to generate the code that the student would write to solve the problem.\newline
Follow these rules carefully:\newline
1. Output only one complete method or function in Java. If a student is unlikely to have knowledge of KCs, the generated code may be incomplete.\newline
2. Do not include any explanations, comments, or formatting symbols (such as backticks).\newline
3. Correctness is not required. The goal is to produce plausible student code influenced by the provided knowledges. Inefficiency, minor bugs, and convoluted logic are expected if the student is unlikely to have knowledge of KCs. This rule is the most important!!!!\newline
The Java coding problem is:<question>\newline
The KC 1 is: <kc1>.\newline
The student has knowledge of <kc1>: <|placeholder|>.\newline
The KC 2 is: <kc2>.\newline
The student has knowledge of <kc2>: <|placeholder|>.\newline
$\dots$\newline
The KC $K$ is: <kcK>.\newline
The student has knowledge of <kcK>: <|placeholder|>.\newline
The code that the student writes to solve this problem based on their knowledge of these KCs is:\\
\bottomrule
\caption{Prompt for the GIRT model.}
\label{tab:gir_model_prompt} \\
\end{longtable}

\begin{longtable}{p{15.0cm}}
\toprule
System:\newline
You are a teacher scoring student written code to a java coding question.\newline\newline
User:\newline
Score the student written code to the java coding question by only outputting a single integer between 0 and 1 inclusive.\newline
Question: <question>\newline
Student written code: <code>\newline
Score:\\
\bottomrule
\caption{Prompt for the scoring model.}
\label{tab:scoring_prompt} \\
\end{longtable}

\end{document}